\documentclass[journal]{IEEEtran}
\ifCLASSINFOpdf
   \usepackage[pdftex]{graphicx}
\else
\fi
\usepackage[cmex10]{amsmath}
\usepackage{algorithmic}
\usepackage{algorithm}

\usepackage{array}
\begin{document}
%
\title{Generalized FMD Detection for Spectrum Sensing under Low Signal-to-Noise Ratio}
%
%
%

\author{Feng Lin,~\IEEEmembership{Student Member,~IEEE,}
        Robert C. Qiu,~\IEEEmembership{Senior Member,~IEEE,}
        Zhen Hu, Shujie Hou,~\IEEEmembership{Student Member,~IEEE,}
				James P. Browning,~\IEEEmembership{Member,~IEEE,}
        Michael C. Wicks,~\IEEEmembership{Fellow,~IEEE}
\thanks{F. Lin, R. C. Qiu, Z. Hu and S. Hou are with the Department
of Electrical and Computer Engineering, Center for Manufacturing Research, Tennessee Technological University, Cookeville, TN, 38505,
e-mail: fenglin@ieee.org; rqiu@tntech.edu; zhu@tntech.edu; shou42@students.tntech.edu}
\thanks{J. P. Browning is with Air Force Research Laboratory, Wright-Patterson AFB, Dayton, OH 45433, e-mail: James.Browning@wpafb.af.mil and M. C. Wicks is with Sensor Systems Division, University of Dayton Research Institute, Dayton, OH 45469, e-mail: Michael.Wicks@udri.udayton.edu.}
}

\maketitle

\begin{abstract}
Spectrum sensing is a fundamental problem in cognitive radio.
We propose a function of covariance matrix based detection algorithm for spectrum sensing in cognitive radio network. Monotonically increasing property of function of matrix involving trace operation is utilized as the cornerstone for this algorithm. The advantage of proposed algorithm is it works under extremely low signal-to-noise ratio, like lower than -30 dB with limited sample data. Theoretical analysis of threshold setting for the algorithm is discussed. A performance comparison between the proposed algorithm and other state-of-the-art methods is provided, by the simulation on captured digital television (DTV) signal. 
\end{abstract}


%
\IEEEpeerreviewmaketitle

\section{Introduction}
%
%
%
%
\IEEEPARstart{A}{s} a limited nature resource, wireless spectrum becomes increasingly scarce due to the evolution of various wireless technique. However, it is not utilized efficiently, the current utilization of a licensed spectrum varies from 15\% to 85\%~\cite{force2002spectrum}. And the number is even lower in rural area. Cognitive radio (CR) is a key technology to mitigate the overcrowding of spectrum space based on its capability of dynamic spectrum access (DSA). 

One fundamental requirement of CR system is to find the frequency hole by secondary user (SU). Therefore, spectrum sensing plays an important role in CR system. Each SU should be able to sense primary user's (PU) existence accurately in low signal-to-noise ratio (SNR) to avoid interference. In some sense, spectrum sensing can be treated as signal detection problem. Generally, matched filter detection,
energy detection, 
cyclostationary feature detection, 
and covariance matrix based detection~\cite{Zeng2007,5089517,4698617,4221495} are used for spectrum sensing. Covariance matrix related spectrum sensing algorithms were extended by employing multiple antennas at the cognitive receiver~\cite{Wang2010}. Feature template matching (FTM)~\cite{Zhanga} extracts signal feature as the leading eigenvector of signal's covariance matrix. 
Kernel principal component analysis (PCA) and kernel generalized likelihood ratio test (GLRT)~\cite{Hou2011} based on matched subspace model map original space data into a higher dimensional feature space. 

However, it is difficult to solve the sensing problem under extremely low SNR, like -30 dB. To circumvent this difficulty, function of matrix based detection (FMD) algorithm will be employed first time for spectrum sensing, which does not require the prior knowledge of structure of either original signal or noise.
If PU signal is included in the received signal, the trace operation on function of received covariance matrix will always be larger than the function of pure noise covariance matrix. Thus we can detect the existence of PU.

\section{System Model}
\label{sys}
In this paper, we consider there is one receive antenna to detect one PU signal. Let $x(t)$ be the continuous-time received signal after unknown channel. Let $T_{s}=1/f_{s}$ be the sampling period, the received signal sample is $x\left [ n \right ]=x\left ( nT_{s} \right )$. There are two hypotheses to detect PU signal's existence, $\mathcal{H}_{0}$, signal does not exist; and $\mathcal{H}_{1}$, signal exists. The received signal samples under the two hypotheses are given respectively as follows:
\begin{equation}
\label{H_0}
\mathcal{H}_{0}:\! x\left [ n \right ]=w\left [ n \right ]
\end{equation}
\begin{equation}
\label{H_1}
\mathcal{H}_{1}:\! x\left [ n \right ]=s\left [ n \right ] + w\left [ n \right ]
\end{equation}
where $w\left [ n \right ]$ is the received white noise, and each sample of $w\left [ n \right ]$ is assumed to be independent identical distribution (i.i.d.), with zero mean and variance $\sigma _{n}^{2}$. $s\left [ n \right ]$ is the received PU signal samples after unknown channel with unknown signal distribution. And, it is assumed that noise and signal are uncorrelated.


Assume spectrum sensing is performed based on the statistics of the $i^{th}$ sensing segment $\Gamma _{x,i}$, which consists of $N_{s}$ sensing vectors with $L$ (called ``smoothing factor") consecutive output samples in each vector:
\begin{equation}
\label{segment}
\Gamma _{x,i}=\left \{ \mathbf{x}_{\left ( i-1 \right )N_{s}+1},\mathbf{x}_{\left ( i-1 \right )N_{s}+2},\cdots ,\mathbf{x}_{\left ( i-1 \right )N_{s}+N_{s}} \right \}
\end{equation}
\begin{equation}
\label{vector_r}
\mathbf{x}_{i}=\left [ x\left [ i \right ],x\left [ i+1 \right ],\cdots ,x\left [ i+L-1 \right ] \right ]^{T}
\end{equation}
where $\mathbf{x}_{i}\sim \mathcal{N}\left ( 0, \mathbf{R}_{x}\right )$. If $N_{s}$ is large enough, $\mathbf{R}_{x}$ can be approximated by sample covariance matrix $\hat{\mathbf{R}}_{x}$:
\begin{equation}
\label{cov_mat}
\hat{\mathbf{R}}_{x}=\frac{1}{N_{s}}\sum_{i=1}^{N_{s}}\mathbf{x}_{i}\mathbf{x}_{i}^{T}
\end{equation}
We will use $\mathbf{R}_{x}$ instead of $\hat{\mathbf{R}}_{x}$ for convenience. Accordingly, we have $\mathbf{R}_{s}$ for $\mathbf{s}_{i}$ and $\mathbf{R}_{n}$ for $\mathbf{w}_{i}$. As we know, $\mathbf{R}_{s}$ is a positive semi-definite matrix with low rank, while $\mathbf{R}_{n}$ is a positive definite matrix. If received signal contains original signal and noise, then we have:
\begin{equation}
\label{cov_mat_equ2}
\mathbf{R}_{x}=\mathbf{R}_{s}+\mathbf{R}_{n}+\mathrm{E}\left ( \mathbf{sw}^{T} \right )+\mathrm{E}\left ( \mathbf{ws}^{T} \right )
\end{equation} 
Since the data samples employed is large enough, we assume signal and noise to be uncorrelated, which implies $\textrm{E}\left ( sw^{T} \right ) = \textrm{E}\left ( ws^{T} \right )=0$, then
\begin{equation}
\label{cov_mat_equ}
\mathbf{R}_{x}=\mathbf{R}_{s}+\mathbf{R}_{n}
\end{equation}
So our two hypotheses can be rewritten as
\begin{equation}
\label{H2_0}
\mathcal{H}_{0}:\! \mathbf{R}_{x}=\mathbf{R}_{n}
\end{equation}
\begin{equation}
\label{H2_1}
\mathcal{H}_{1}:\! \mathbf{R}_{x}=\mathbf{R}_{s}+\mathbf{R}_{n}
\end{equation}
The detection will be based upon the covariance matrix of received signal in sensing segment $\Gamma _{x,i}$.


\section{Function of Matrix Based Detection}
\label{dete}

\subsection{Spectrum Sensing Algorithm}
The total received sensing segment $\Gamma _{x,i}$ is divided into $K$ sub-segments, with $N_{k}$ vectors in each sub-segment, where $N_{k}= N_{s}/K$. $\Gamma _{x,i}$ can be represented as
\begin{equation}
\label{subsegment}
\Gamma _{x,i}=\left \{ \varphi_{1},\varphi_{2},\cdots,\varphi_{K}   \right \}
\end{equation}
The $k^{th}$ sub-segment $\gamma_{k}$ is represented as
\begin{equation}
\label{subsegment1}
\varphi_{k}=\left \{ \mathbf{x}_{\left ( k-1 \right )N_{k}+1},\mathbf{x}_{\left ( k-1 \right )N_{k}+2},\cdots ,\mathbf{x}_{\left ( k-1 \right )N_{k}+N_{k}} \right \}
\end{equation}
Then we calculate the sample covariance matrix from each sensing sub-segment separately, using Eq.~(\ref{cov_mat}) with sum interval from 1 to $N_{s}$ replaced by different sub-segment interval. Covariance matrix of $k^{th}$ sub-segment is denoted as 
\begin{equation}
\label{cov_mat_sub}
\mathbf{R}_{x,k}=\frac{1}{N_{k}}\sum_{i=(k-1)N_{k}+1}^{(k-1)N_{k}+N_{k}}\mathbf{x}_{i}\mathbf{x}_{i}^{T}
\end{equation}
Accordingly, we have original signal covariance matrix $\mathbf{R}_{s,k}$ and noise covariance matrix $\mathbf{R}_{n,k}$. Similar to Eq.~(\ref{cov_mat_equ}), if original signal is contained, we have
\begin{equation}
\label{cov_mat_equ_sub}
\mathbf{R}_{x,k}=\mathbf{R}_{s,k}+\mathbf{R}_{n,k}
\end{equation}
Because $\mathbf{R}_{s,k}$ is positive semi-definite matrix, for all $k$
\begin{equation}
\label{mat_larger}
\mathbf{R}_{s,k}+\mathbf{R}_{n,k}\succ \mathbf{R}_{n,k}
\end{equation}


In order to make the metric robust, we calculate the sample mean of the covariance matrix to get a average covariance matrix. From
Eq.~(\ref{mat_larger}), we get 
\begin{equation}
\label{mat_larger_ave}
\frac{1}{K}\sum_{k=1}^{K}\left (\mathbf{R}_{s,k}+\mathbf{R}_{n,k}  \right )\succ \frac{1}{K}\sum_{k=1}^{K}\mathbf{R}_{n,k}
\end{equation}

Generally, if $\mathbf{A}\prec  \mathbf{B}$, we may find many appropriate monotonically increasing functions applied to both sides of the matrix inequality with matrix inequality holds that $f\left ( \mathbf{A} \right )\prec f\left ( \mathbf{B} \right )$, as function of matrix is still matrix. But not all the monotonically increasing functions satisfy $f\left ( \mathbf{A} \right )\prec f\left ( \mathbf{B} \right )$.
Fortunately, if we apply $\textrm{trace}$ operation to both sides of the function $f$, the real number inequality always holds, as shown in \textbf{Fact 1.}~\cite[p.~529]{matrixmath}.

\textbf{Fact 1.} Let $\mathbf{A}, \mathbf{B}\in \textrm{F}^{n\times n}$ ($\textrm{F}^{n}$ means $\textrm{R}^{n}$ or $\textrm{C}^{n}$), assume that $\mathbf{A}$ and $\mathbf{B}$ are positive semidefinite, assume that $\mathbf{A}\preceq  \mathbf{B}$, let $f:[0,\infty) \mapsto [0,\infty)$, and assume that $f(0)=0$, $f$ is continuous, and $f$ is increasing. Then,
\begin{equation}
\label{func_Tr}
\textrm{Tr}\left ( f\left ( \mathbf{A} \right ) \right )\leq \textrm{Tr}\left ( f\left ( \mathbf{B} \right ) \right )
\end{equation}
Based on Eq.~(\ref{mat_larger_ave}) and (\ref{func_Tr}), we can get
\begin{equation}
\label{mono_sum_f_Tr}
\begin{array}{l}
 {\rm Tr}\left( {f\left( {\frac{1}{K}\sum\limits_{k = 1}^K {\left( {{\bf R}_{s,k}  + {\bf R}_{n,k} } \right)} } \right)} \right) \\ 
 {\rm  > Tr}\left( {f\left( {\frac{1}{K}\sum\limits_{k = 1}^K {{\bf R}_{n,k} } } \right)} \right) \\ 
\end{array}
\end{equation}
We can see from the above results that if PU signal exists, the left side of Eq.~(\ref{mono_sum_f_Tr}) will be larger than right side. Because we have averaged the covariance matrices of sub-segments, the numerical difference between them is stable. As a result, we can find a threshold to distinguish the PU signal present or absent.

Based on the analysis above, we propose detection algorithm 1 for spectrum sensing. Namely, FMD.
\algsetup{indent=2em}
\begin{algorithm}
\caption{FMD}
\label{alg_single}
\begin{algorithmic}[1]

\STATE Determine probability of false alarm $P_{fa}$ and threshold $\gamma$.
\STATE Divide the total received segment into $K$ sub-segments, and calculate their covariance matrices.
\STATE Get an average covariance matrix from all covariance matrices of sub-segments
\STATE Find an appropriate rational function $f$ and get the function $f$ result of average covariance matrix.
\STATE Obtain the trace of the function of covariance matrix as a metric $\rho$
\begin{equation}
\label{rho}
\rho ={\rm Tr}\left (f\left (\frac{1}{K}\sum_{k=1}^{K}\mathbf{R}_{x,k}  \right )  \right )
\end{equation}

\IF{$\rho > \gamma$}
	\STATE PU signal exists
\ELSE
  \STATE PU signal does not exist
\ENDIF

\end{algorithmic}
\end{algorithm}

\subsection{Detection and Threshold Setting}
\label{threshold}
The optimal choice of appropriate monotonically increasing function is data dependent. For simplicity, we choose function $f(\mathbf{A})=\mathbf{A}$, which is monotonically increasing and can achieve near best performance among all the potential functions we experienced. Thus, the metric Eq.~(\ref{rho}) can be specified to
\begin{equation}
\label{rho_sim}
\rho_{1} =\mathrm{Tr}\left (\frac{1}{K}\sum_{k=1}^{K}\mathbf{R}_{x,k} \right )
\end{equation}

In Eq.~(\ref{rho_sim}), all the covariance matrices are added together firstly, then Tr operation is applied. 
Since Tr operation is nothing but a sum operation of diagonal elements of covariance matrix. We can apply Tr operation to covariance matrix firstly, then add them together to get the same $\rho_{1}$, which can be represented as
\begin{equation}
\label{rho_sim2}
\rho_{1} =\frac{1}{K}\sum_{k=1}^{K}\mathrm{Tr}\left (\mathbf{R}_{x,k} \right )
\end{equation}
In this paper, we use Eq.~(\ref{rho_sim2}) to determine probability of false alarm and the threshold.

For hypothesis $\mathcal{H}_{0}$, original signal is not contained in received signal. Since the number of vectors to calculate covariance matrix $N_{k}$ is very large, based on Central Limit Theorem, $\frac{1}{L}\mathrm{Tr}\left ( \mathbf{R}_{x} \right )$ can be approximated by Gaussian distribution with mean $\sigma _{n}^{2}$ and variance $\frac{2\sigma _{n}^{4}}{N_{k}}$~\cite{4221495}. Then
\begin{equation}
\label{Tr_dis}
\mathrm{Tr}\left ( \mathbf{R}_{x,k} \right )\sim \mathcal{N}\left ( L\sigma_{n} ^{2},\frac{2L^{2}}{N_{k}}\sigma _{n}^{4} \right )
\end{equation}
\begin{equation}
\label{ETr_dis}
\rho_{1} =\frac{1}{K}\sum_{k=1}^{K}\mathrm{Tr}\left ( \mathbf{R}_{x,k} \right )\sim \mathcal{N}\left ( L\sigma_{n} ^{2},\frac{2L^{2}}{KN_{k}}\sigma _{n}^{4} \right )
\end{equation}

The variance of noise $\sigma _{n}^{2}$ can be estimated from Eq.~(\ref{Tr_dis}).
\begin{equation}
\label{sigma2}
\hat{\sigma }_{n}^{2}= \sqrt{\frac{N_{k}}{2L^{2}}\hat{\sigma}^{2} }
\end{equation} 
$\hat{\sigma}^{2}$ is the sample variance of $\mathrm{Tr}\left ( \mathbf{R}_{x,k} \right )$, which can be obtained from unbiased estimator:
\begin{equation}
\label{sample_var}
\hat{\sigma}^{2}=\frac{1}{K-1}\sum_{k=1}^{K}\left ( Z_{k} -\bar{Z}_{K}\right )^{2}
\end{equation}
Where,
\begin{equation}
\label{Z}
Z_{k}=\mathrm{Tr}\left ( \mathbf{R}_{x,k} \right )
\end{equation}
\begin{equation}
\label{Z2}
\bar{Z}_{K}=\frac{1}{K}\sum_{k=1}^{K}\mathrm{Tr}\left ( \mathbf{R}_{x,k} \right )
\end{equation}
Then $\sigma_{n} ^{2}$ is replaced by $\hat{\sigma }_{n}^{2}$ in the Eq.~(\ref{ETr_dis}) to be
\begin{equation}
\label{ETr_dis2}
\rho_{1} \sim \mathcal{N}\left ( L\hat{\sigma_{n}} ^{2},\frac{2L^{2}}{KN_{k}}\hat{\sigma} _{n}^{4} \right )
\end{equation}

The choice of threshold $\gamma$ is a compromise between $P_{d}$ and $P_{fa}$. Usually, it is difficult to set threshold based on $P_{d}$, so we choose the threshold based on $P_{fa}$. The probability of false alarm for FMD is
\begin{equation}
\label{Pfa}
P_{fa}=P\left ( \rho_{1}>\gamma \mid \mathcal{H}_{0} \right )
\end{equation}
\begin{equation}
\label{Pfa2}
P_{fa}=P\left ( \frac{\rho_{1}-L\hat{\sigma_{n}} ^{2}}{\sqrt{\frac{2L^{2}}{KN_{k}}\hat{\sigma_{n}} ^{4}}}> \frac{\gamma -L\hat{\sigma_{n}} ^{2}}{\sqrt{\frac{2L^{2}}{KN_{k}}\hat{\sigma_{n}} ^{4}}} \right )
\end{equation}
\begin{equation}
\label{Pfa3}
P_{fa}=Q\left (\frac{\gamma -L\hat{\sigma_{n}} ^{2}}{\sqrt{\frac{2L^{2}}{KN_{k}}\hat{\sigma_{n}} ^{4}}} \right )
\end{equation}
where
\begin{equation}
\label{Q}
Q\left ( t \right )=\frac{1}{\sqrt{2\pi }}\int_{t}^{+\infty }e^{\frac{-x^{2}}{2}}\mathrm{d}x
\end{equation}
Then we get the threshold $\gamma$
\begin{equation}
\label{gamma_f}
\gamma = \left (1+Q^{-1}\left ( P_{fa} \right )\sqrt{\frac{2}{KN_{k}}} \right )L\hat{\sigma_{n}} ^{2}
\end{equation}


\section{Covariance Matrix Based Algorithms}
\label{cov}
Several sample covariance matrix based algorithms have been proposed in spectrum sensing. Maximum-minimum eigenvalue (MME)~\cite{Zeng2007} and arithmetic-to-geometric mean (AGM)~\cite{4698617} have no prior knowledge, while FTM~\cite{Zhanga} has feature as prior knowledge. Estimator-Correlator (EC) requires both original signal covariance matrix and noise variance.
All the threshold $\gamma$ are determined by probability of false alarm if not specified.

\subsection{MME}
The algorithm first calculates the maximum and minimum eigenvalue of sample covariance matrix, namely, $\lambda _{max}$ and $\lambda _{min}$. If there is no signal, $\lambda _{max}/\lambda _{min}=1$; otherwise, $\lambda _{max}/\lambda _{min}>1$. So the resulting detector computes the ratio of $\lambda _{max}$ and $\lambda _{min}$ and is compared with a threshold
\begin{equation}
\label{mme}
T_{MME}=\frac{\lambda _{max}}{\lambda _{min}}> \gamma _{MME} 
\end{equation}

\subsection{AGM}
AGM is derived from GLRT, it finds an unstructured estimate of $\mathbf{R}_{x}$ to be $\mathbf{R}_{s}+\sigma_{n}^{2}\mathbf{I}$ under $\mathcal{H}_{1}$ and $\sigma_{n}^{2}\mathbf{I}$ under $\mathcal{H}_{0}$. $\lambda_{m}$ represent all eigenvalues of sample covariance matrix and $M$ is the dimension of sample covariance matrix. Since the arithmetic mean is larger than geometric mean, the resulting detector computes the AGM of the eigenvalues of sample covariance matrix and compares with a threshold
\begin{equation}
\label{AGM}
T_{AGM}= \frac{\frac{1}{M}\sum_{m}\lambda_{m}}{\left (\prod_{m}\lambda_{m}  \right )^{\frac{1}{M}} }> \gamma _{AGM}
\end{equation}

\subsection{FTM}
FTM extracts leading eigenvector as the feature $\eta _i$, which is stable for signals and random for noise.  
First, it learns the feature blindly by comparing similarity of two consecutive sensing segment, namely feature learning algorithm (FLA).
\begin{equation}
\label{FLA}
T_{FLA}  = \mathop {\max }\limits_{l = 1,2, \cdots L - k + 1} \left| {\sum\limits_{k = 1}^L {\eta _i \left[ k \right]\eta _{i + 1} \left[ {k + l} \right]} } \right|
\end{equation}
If $T_{FLA} > \gamma_{e}$ feature is learned as $\phi _{s,1}=\eta _{i+1}$.
 Then with the learned signal feature $\phi _{s,1}$ as prior knowledge, this algorithm just compares the feature $\phi _{x,1}$ from new sensing segment and signal feature $\phi _{s,1}$ to determine if signal exists.
\begin{equation}
\label{FTM}
T_{FTM}  = \mathop {\max }\limits_{l = 1,2, \cdots L - k + 1} \left| {\sum\limits_{k = 1}^L {\phi _{s,1} \left[ k \right]\phi _{x,1} \left[ {k + l} \right]} } \right| > \gamma _{FTM} 
\end{equation}

\subsection{EC}
EC~\cite{dete_theo} method assumes the signal follows zero mean Gaussian distribution with covariance matrix $\mathbf{R}_{s}$, and noise follows zero mean Gaussian distribution with covariance matrix $\sigma _{n}^{2}\mathbf{I}$,
\begin{equation}
\label{EC1}
s\sim \mathcal{N}\left ( 0,\mathbf{R}_{s} \right ),n\sim \mathcal{N}\left ( 0,\sigma _{n}^{2}\mathbf{I} \right )
\end{equation}
Both $\mathbf{R}_{s}$ and $\sigma_{n}^{2}$ are given priorly. The hypothesis $\mathcal{H}_{1}$ is true if:
\begin{equation}
\label{EC2}
T_{EC}= \sum_{j=1}^{N_{s}}\mathbf{x}_{j}^{T}\mathbf{R}_{s}\left (\mathbf{R}_{s}+\sigma _{n}^{2}\mathbf{I } \right )^{-1}\mathbf{x}_{j}>\gamma _{EC}
\end{equation}

\subsection{Computational Complexity}
The computation complexity comes from two parts: sample covariance matrix calculation and eigenvalue decomposition of covariance matrix. Besides $LN_{s}$ multiplications and additions for MME, AGM, FTM, EC to calculate sample covariance matrix, FMD requires $K$ additional additions to calculate covariance matrix. Generally $\mathcal{O}\left ( L^{3} \right )$ multiplications and additions are sufficient for MME, AGM, FTM to calculate eigenvalue decomposition. While FMD needs 1 function of matrix calculation, computational complexity depends on what kind of function is chosen.
\section{Simulation Results}
\label{simu}
In the following, we will give some simulation results using
DTV signal (field measurements) captured in Washington D.C.~\cite{DTV2006Measurements}. The data rate of the vestigial sideband (VSB) DTV signal is 10.762 MSamples/sec. The recorded DTV signals were sampled at receiver at 21.524476 MSamples/sec and down converted to a low central IF frequency of 5.381119 MHz.
The number of total samples contained in the segment is $N_{s}=100,000$ (corresponding to about 5ms sampling time), and they are divided into 166 sub-segments with 600 samples in each sub-segment. The smoothing factor $L$ is chosen to be 32. Probability of false alarm is fixed with $P_{fa}=1\%$. The SNR of the received signal are unknown. In order to use the signals for simulating the algorithms at very low SNR, we need to add white Gaussian noise to obtain various SNR levels. 2,000 simulations are performed on each SNR level.

\begin{figure}[!t]
\centering
\includegraphics[width=3.4in]{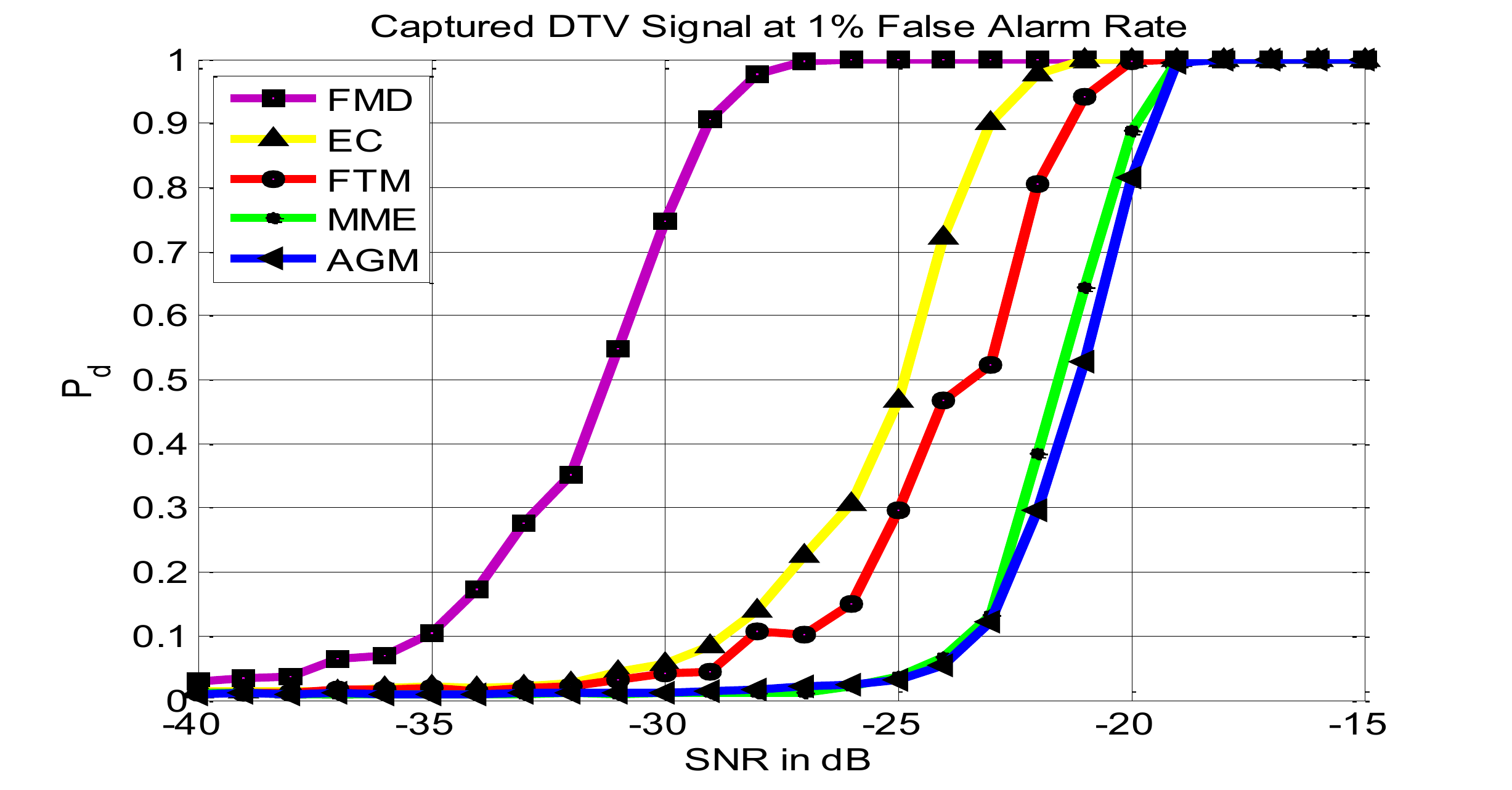}
\caption{Probability of detection at 1\% probability of false alarm}
\label{FMD_compare_dtv}
\end{figure}

The probabilities of detection varied by SNR for FMD compared with EC, AGM, FTM, MME are shown in Fig.~\ref{FMD_compare_dtv}. FMD method is better than all other methods introduced in Section~\ref{cov}, which can achieve 0.5 probability of detection at SNR -31.5 dB. In addition, The performance can be improved further if we use more data samples. We can see from Table~\ref{table_FMD_EC}, the SNR of FMD can even achieve -34 dB at 0.5 probability of detection employing 2e5 data. Meanwhile, the tendency of gain between FMD and EC is apparent that it is increasing with more data samples. This is because more samples can make the numerical difference between two hypotheses more stable, hence the threshold can totally separate signals from two hypotheses. 


\begin{table}[!t]
\renewcommand{\arraystretch}{1.3}
\caption{Gain between FMD and EC using different number of data}
\label{table_FMD_EC}
\centering
\begin{tabular}{|c||c||c||c||c||c|}
\hline
Number of samples  & 2.5e4 & 5e4 & 1e5 & 2e5\\
\hline
SNR of FMD  & -26 dB & -30 dB & -31.5 dB & -34 dB\\
\hline
SNR of EC  & -22 dB & -23.5 dB & -25 dB & -26.5 dB\\
\hline
Gain  & 4 dB & 6.5 dB & 6.5 dB & 7.5 dB\\
\hline
\end{tabular}
\end{table}

\section{Conclusion}
\label{conc}
In this paper we considered the spectrum sensing for single PU with single antenna, and FMD algorithm is proposed.  
Captured DTV signal is simulated with our proposed algorithm and other state-of-the-art algorithms, including MME, FTM, AGM. The simulation results showed our proposed algorithm can work lower than -30 dB SNR with limited data, which can even beat the performance of EC with perfect prior knowledge. 


The optimal number for each subsegment of 100,000 total data is 600. We need to find out how to optimally choose the number of data to calculate covariance matrix under different total sample data in the future.
%



%
%


%

%

\section*{Acknowledgment}
This work is funded by National Science Foundation through two grants (ECCS-0901420 and ECCS-0821658), and
Office of Naval Research through two grants (N00010-10-1-0810 and N00014-11-1-0006). 

\ifCLASSOPTIONcaptionsoff
  \newpage
\fi


\bibliographystyle{IEEEtran}
\bibliography{bib/cognitive_radio_zhe,bib/library,bib/CR_Peng,bib/Pattern_Recognition,bib/Eric_ICNC2012,bib/Qiu_Group_bib,bib/Matrix_Inequality2}
\end{document}